\newif\ifmydraft\mydraftfalse
\newif\ifmyanon\myanonfalse
\newif\ifmyarxiv\myarxivtrue
\documentclass[sigplan,screen\ifmydraft,review\fi\ifmyarxiv,nonacm\fi\ifmyanon,anonymous\fi]{acmart}
\AtBeginDocument{%
  }

\setcopyright{acmlicensed}
\copyrightyear{2018}
\acmYear{2018}
\acmDOI{XXXXXXX.XXXXXXX}
\acmConference[Conference acronym 'XX]{Make sure to enter the correct
  conference title from your rights confirmation email}{June 03--05,
  2018}{Woodstock, NY}
\acmISBN{978-1-4503-XXXX-X/2018/06}

\usepackage{xcolor}
\usepackage{xspace}
\usepackage{listings}
\usepackage[most]{tcolorbox}
\usepackage{lipsum} %
\usepackage{caption}
\usepackage{enumitem}
\usepackage{subcaption}

\newtcolorbox{userbox}{
  colback=blue!5!white,
  colframe=blue!75!black,
  fonttitle=\bfseries,
  title=Logic Programming Interpreter,
  boxrule=0.5pt,
  arc=2pt,
  left=4pt,
  right=4pt,
  top=4pt,
  bottom=4pt
}

\newtcolorbox{prompt}{
  colback=blue!5!white,
  colframe=blue!75!black,
  fonttitle=\bfseries,
  boxrule=0.5pt,
  arc=2pt,
  left=4pt,
  right=4pt,
  top=4pt,
  bottom=4pt
}

\newtcolorbox{llmbox}{
  colback=gray!10!white,
  colframe=black!70,
  fonttitle=\bfseries,
  title=LLM,
  boxrule=0.5pt,
  arc=2pt,
  left=4pt,
  right=4pt,
  top=4pt,
  bottom=4pt
}

\definecolor{commentgray}{gray}{0.5}
\definecolor{stringpurple}{rgb}{0.58,0,0.82}
\definecolor{keywordblue}{rgb}{0,0,0.7}

\lstdefinelanguage{Prolog}{
    morekeywords={is,not,mod,div,fail,true,false,consult,assert,retract,rule,fact,clause,listing,once},
    sensitive=true,
    morecomment=[l]\%,
    morestring=[b]',
}

\newcommand{\nesy}{NeSy\xspace}

\newcommand{\nsdm}{NSTS\xspace}

\definecolor{neuralblue}{RGB}{118,214,255}
\definecolor{symbolicorange}{RGB}{255,212,121}

\newcommand{\hl}[2][yellow]{{\setlength{\fboxsep}{0pt}{\colorbox{#1}{#2}}}}

\begin{document}

\newcommand{\mytitle}{Current Practices for Building LLM-Powered Reasoning Tools Are Ad Hoc---and We Can Do Better}
\ifmyarxiv
\title{\mytitle}
\else
\title[We Can Do Better Than the Current Ad Hoc Practices for Building LLM-Powered AR Tools]{\mytitle}
\fi

\author{Aaron Bembenek}
\email{aaron.bembenek@unimelb.edu.au}
\orcid{0000-0002-3677-701X}
\affiliation{%
  \institution{School of Computing and Information Systems, The University of Melbourne}
  \city{Parkville}
  \state{Victoria}
  \country{Australia}
}

\begin{abstract}
  
There is growing excitement about building software verifiers, synthesizers, and other Automated Reasoning (AR) tools by combining traditional symbolic algorithms and Large Language Models (LLMs).
Unfortunately, the current practice for constructing such neurosymbolic AR systems is an ad hoc programming model that does not have the strong guarantees of traditional symbolic algorithms, nor a deep enough synchronization of neural networks and symbolic reasoning to unlock the full potential of LLM-powered reasoning.
I propose Neurosymbolic Transition Systems as a principled computational model that can underlie infrastructure for building neurosymbolic AR tools.
In this model, symbolic state is paired with intuition, and state transitions operate over symbols and intuition in parallel.
I argue why this new paradigm can scale logical reasoning beyond current capabilities while retaining the strong guarantees of symbolic algorithms, and I sketch out how the computational model I propose can be reified in a logic programming language.

\end{abstract}


\received{20 February 2007}
\received[revised]{12 March 2009}
\received[accepted]{5 June 2009}

\maketitle

\section{Introduction}

Traditionally, Automated Reasoning (AR) tools---systems designed for explicit logical reasoning, including automated theorem provers, software verifiers, and synthesizers---have been implemented as symbolic algorithms.
These algorithms systematically explore the solution space and need good heuristics to scale to larger problems.
Given the effectiveness of learned heuristics in guiding search~\cite{Irving2016DeepMath,Jakubuv2017ENIGMA,Loos2017Deep,Zhang2018Neural,Kalyan2018Neural,Selsam2019Guiding}, there has been growing interest in neurosymbolic (\nesy) AR tools that integrate symbolic algorithms with Large Language Models (LLMs).
The hope is that LLMs can use their vast implicit knowledge and generative capabilities to overcome the scalability limits of pure symbolic reasoning, without necessitating custom training for each new AR task.

While there is much to gain from building AR tools integrating symbolic algorithms and LLMs, the two are very different types of computations, and it is not clear how to combine them in a way that brings out the full potential of both paradigms, while ensuring that the overall AR tool has desirable properties.
It is also an open question whether programming language infrastructure can be designed to help developers build \nesy AR tools in a principled way.

Given the current lack of systematized principles for building LLM-powered AR tools, developers implicitly adopt what I characterize as the {\em sequential architecture for \nesy systems}---a natural, but (I argue) suboptimal approach to fusing LLMs and symbolic reasoning.
In this model, LLM calls and symbolic algorithms are distinct components chained {\em in sequence} and separated by implicit neural-symbolic boundaries, with the result that, at any one time, a piece of data is {\em exclusively neural or symbolic}, depending on which side of the boundary it is on.
Strict neural-symbolic boundaries preclude some benefits of neurosymbolic programming.
Moreover, this architecture forces computation to flow through illogical components (LLMs, which cannot guarantee correct reasoning), making it an awkward foundation for logical AR tools.

I propose Neurosymbolic Transition Systems ({\nsdm}s) as an alternative computational model for \nesy AR tools.
In this model, a developer writes a symbolic algorithm, and the \nsdm runtime lifts it to a \nesy algorithm that retains key properties of the original computation (e.g., soundness, progress).
The \nsdm treats the states and state transitions of the underlying symbolic algorithm as being {\em functionally neurosymbolic}; it does so by augmenting symbolic components with intuition
(where ``intuition'' abstractly refers to data relevant to neural inference).
The \nsdm maintains a dual state pairing a symbolic state---tracking logical structure---with intuition that captures context.
During computation, symbolic state and intuition evolve {\em in parallel}, as the \nsdm updates its current intuition with the intuition associated with the symbolic steps that are taken.
At decision points, the \nsdm uses its accumulated intuition to guide symbolic reasoning, enabling more efficient exploration of the solution space.

To accelerate its impact, the field of LLM-powered AR needs infrastructure for \nesy programming that can be shared across AR systems.
{\nsdm}s---a powerful conceptual model for how to structure \nesy computation---can be the theoretical foundation for this important infrastructure.

\paragraph{Contributions}

In this position paper, I identify the weaknesses of the sequential architecture for \nesy AR tools (Section~\ref{sec:seq}); propose Neurosymbolic Transition Systems as a new \nesy computational model (Section~\ref{sec:par}); and sketch out a logic programming language based on this model (Section~\ref{sec:lang}).

\paragraph{Related Work}

SymbolicAI~\cite{Dinu2024SymbolicAI} and Chain of Code~\cite{Li2024Chain} are frameworks that coordinate LLM calls and symbolic computation within the sequential \nesy architecture; the \nsdm model could provide a methodology for integrating \nesy AR components into these LLM-powered frameworks.
Logic programming has been used to implement advanced AR tools~\cite{Podelski2007ARMC,Jaffar2012TRACER,Grebenshchikov2012HSFC,DeAngelis2014VeriMAP,Kafle2015Tree} and is popular in the \nesy domain~\cite{Manhaeve2021Neural,Huang2021Scallop,Li2024Relational,Vakharia2024ProSLM,Wang2024ChatLogic}.
The \nsdm model builds on work using neural models to guide symbolic algorithms~\cite{Irving2016DeepMath,Jakubuv2017ENIGMA,Loos2017Deep,Zhang2018Neural,Kalyan2018Neural,Selsam2019Guiding}.
The language implementation I propose generalizes the approach of a tool~\citep{Li2024Guiding} that uses an LLM to update a probabilistic model that guides enumerative program synthesis, and updates the LLM with feedback from symbolic search; this tool outperforms purely neural or symbolic approaches.

\section{The Sequential \nesy Architecture}\label{sec:seq}

In the absence of a principled programming model for building \nesy AR systems, it is natural to sequentially chain LLM calls and symbolic algorithms.
I identify two main limitations with this ad hoc approach: this sequential architecture induces implicit neural-symbolic boundaries that make it hard to achieve the full benefits of \nesy programming (Section~\ref{sec:boundary}), and this architecture tends to result in \nesy AR tools with suboptimal computational properties (Section~\ref{sec:guarantees}).

\subsection{Strict Neural-Symbolic Boundaries}\label{sec:boundary}

Sequentially chaining LLM calls and symbolic algorithms induces implicit neural-symbolic boundaries (the dashed line in Figure~\ref{fig:seq_cegis}).
On one side of a boundary is the world of neural intuition; on the other side of that boundary is the world of symbolic reasoning.
When a piece of data crosses a neurosymbolic boundary, it is implicitly reinterpreted in that new world.
This strict separation of worlds precludes some potential benefits of neurosymbolic programming.

First, when a symbolic datum flows into the neural world, there is no guarantee that the LLM interprets it according to the intended semantics it had in the symbolic world.
Thus, any computation performed by the LLM on that datum is not guaranteed to produce an answer consistent with symbolic reasoning.
The \nsdm model is designed to sidestep this limitation, without requiring any changes to the LLM itself.

Second, arbitrary symbolic algorithms programmed in existing languages do not use neural intuition to guide their search for a solution, or produce intuition to explain their solution.
This is a loss.
In the first place, the intuitive rationale of the LLM (i.e., its ``proof intentions''~\cite{Zhang2025Position}) might help the symbolic algorithm more effectively leverage the data it gets from the LLM.
In the second place, the intuition behind a symbolic conclusion could help the LLM more effectively use that conclusion (and would also be helpful for a human analyst).
In the \nsdm model, symbolic algorithms 1) are automatically guided by intuition, and 2) automatically produce intuition for their symbolic conclusions.

\subsection{Suboptimal Computational Properties}\label{sec:guarantees}

\begin{figure}
    \includegraphics[width=\columnwidth]{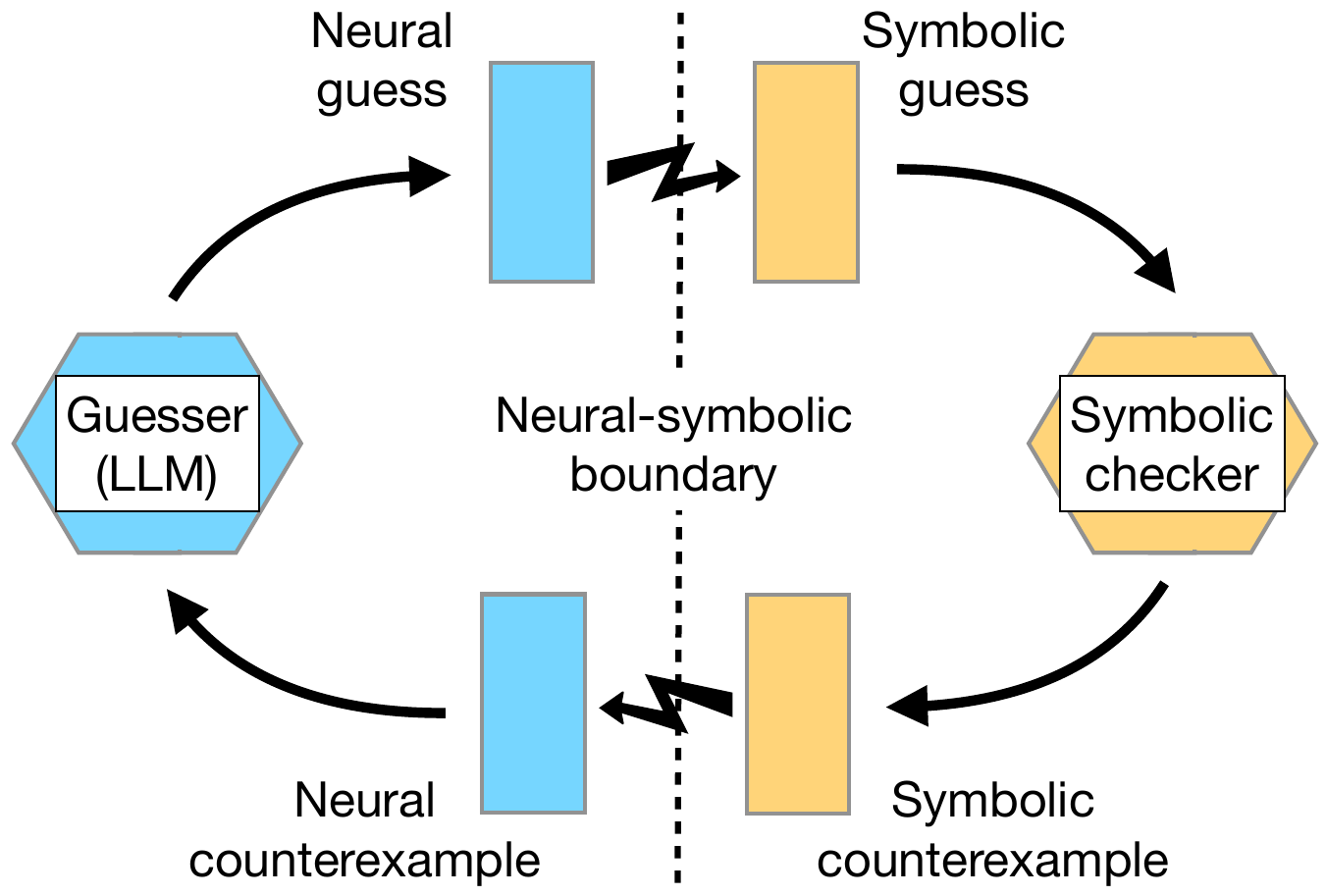}
    \caption{Current LLM-based AR tools have distinct neural and symbolic components, inducing implicit neural-symbolic boundaries (the dashed line); a datum is reinterpreted when it flows across a boundary. Guess-and-check loops like the one here are not guaranteed to make progress. In all figures, \hl[neuralblue]{blue} indicates the neural world, \hl[symbolicorange]{orange} the symbolic world.}\label{fig:seq_cegis}
\end{figure}

Consider a hypothetical LLM-powered SAT solver.
Is it a decision procedure?
What is its computational complexity?
What is the asymptotic growth rate of the number of LLM calls?
The sequential \nesy architecture tends to result in computations with unsatisfactory answers to these questions, undermining its suitability as an AR programming model.

For example, many LLM-powered AR tools~\cite{Kamath2024Leveraging,Wu2024LLM,Jha2023Counterexample,Orvalho2025Counterexample,Pan2023Logic,Wu2024Lemur} contain some form of a \nesy guess-and-check loop (Figure~\ref{fig:seq_cegis}), where the LLM proposes a solution, a symbolic verifier checks it and, if it is not valid, returns feedback to the LLM; the loop repeats until a solution is verified.
It is natural to view these \nesy guess-and-check loops as analogous to the {\em symbolic} guess-and-check loops in core AR algorithms like
CEGAR~\cite{Clarke2000Counterexample}, CEGIS~\cite{SolarLezama2006Combinatorial}, and DPLL(\emph{T})~\cite{Nieuwenhuis2006Solving}.
But, this analogy is inaccurate: unlike a correctly implemented symbolic guesser, there is no guarantee that the LLM will produce candidate solutions that are consistent with the feedback it has received.
In the presence of a chronically hallucinating LLM, the loop will not make progress and the tool will not converge on a solution, even if the solution space is finite (e.g., a SAT problem); plus, the number of LLM calls is unbounded.

In contrast, a computation in the \nsdm model inherits the worst-case complexity of the corresponding symbolic algorithm (assuming constant-time LLM calls).
Moreover, it is often not necessary to sacrifice expressiveness to gain better computational properties; in fact,
it is conceptually straightforward to port a sequential \nesy guess-and-check loop---which might have no completeness or termination guarantees---to an analogous computation in the \nsdm model with better properties (e.g., a semi-decision procedure).

\section{Neurosymbolic Transition Systems}\label{sec:par}

An \nsdm is built on top of a transition system (which I refer to as the {\em symbolic} transition system).
A transition system~\cite{baier2008principles} consists of states and a transition relation between states; it is a powerful model that encompasses common abstractions of program evaluation, like small-step operational semantics~\cite{Plotkin2004Structural} and SECD machines~\cite{Landin1964Mechanical}.
An \nsdm augments a transition system with 1) a way to build intuition about an initial symbolic state, and 2) a way to build intuition for each transition of the symbolic system.
I use ``intuition'' to refer abstractly to data relevant to neural inference.
If $\mathcal{I}$ is the domain of intuition, I assume the existence of an operator $\mathsf{combine} : \mathcal{I} \times \mathcal{I} \rightarrow \mathcal{I}$ for combining two pieces of intuition, and an operator $\mathsf{infer} : \mathcal{I} \rightarrow \mathcal{I}$ for making an informal inference over intuition.
In a naive setting, the domain $\mathcal{I}$ could be text, the operator $\mathsf{combine}$ could be concatenation, and the operator $\mathsf{infer}$ could prompt an LLM with the text.

\begin{figure}
    \centering
    \includegraphics[width=\columnwidth]{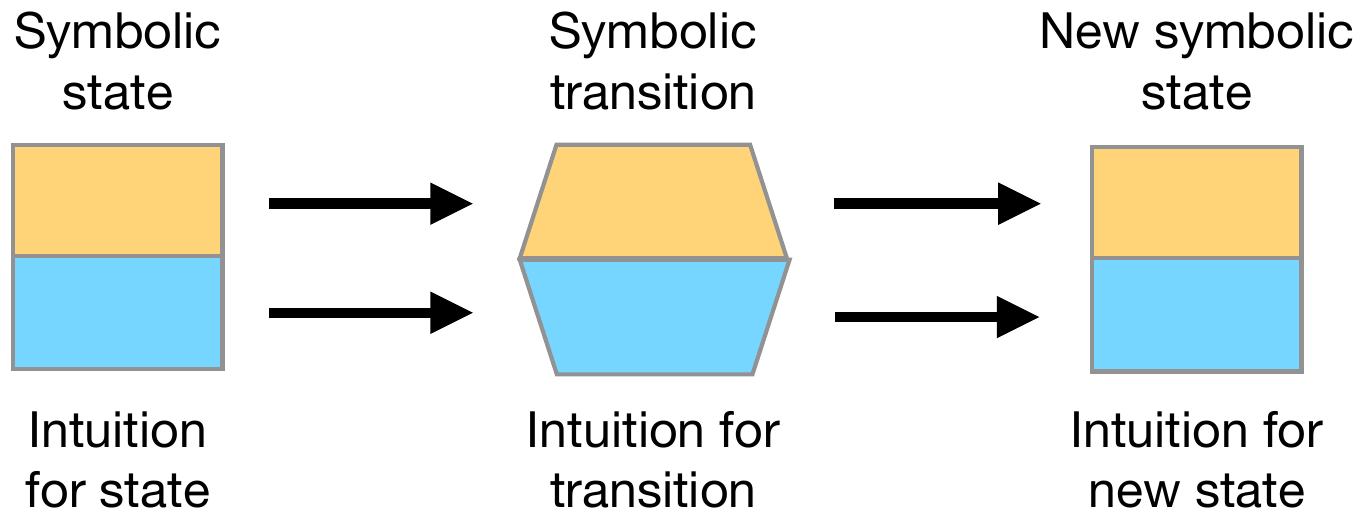}
    \caption{During \nsdm evaluation, symbolic states and symbolic transitions are associated with intuition, and computation is performed over symbols and intuition in parallel.}\label{fig:computation}
\end{figure}

During evaluation, the state of an \nsdm is a pair consisting of a symbolic state and intuition.
The \nsdm steps symbolically and intuitively {\em in parallel} (Figure~\ref{fig:computation}): when the underlying symbolic transition system steps according to its transition relation, the \nsdm updates its intuition, via the $\mathsf{combine}$ operator, with the intuition for that symbolic transition.
When the \nsdm reaches a final state, the aggregated intuition can be used to construct an intuitive summary of the symbolic transitions that have been performed.

An {\nsdm} is most useful if the underlying transition system is non-deterministic---i.e., some states have multiple outgoing transitions. 
When such a transition system is actually implemented, non-deterministic choice is replaced by some mechanism that makes choices (e.g., a heuristic for minimizing the distance to a final state). 
In the case of an \nsdm, the mechanism that makes choices can also get guidance from the current intuition, via the $\mathsf{infer}$ operator.
Built up over the course of evaluation, intuition can be a rich source of information for making decisions, helping the symbolic transition system cut through state explosion (at least in principle).

\paragraph{Guessing and Checking}

LLM-powered AR tools often contain \nesy guess-and-check loops in which a symbolic checker provides counterexamples to an LLM until the LLM guesses correctly (Figure~\ref{fig:seq_cegis}).
In the \nsdm model, programmers write code that is evaluated via a symbolic transition system, and the runtime lifts symbolic evaluation to \nesy evaluation.
As the program does not have access to the \nsdm intuition, it would not contain explicit \nesy guess-and-check loops.
Nonetheless, an {\nsdm} can still perform neural guessing and symbolic checking.
Say an \nsdm uses an LLM to maintain intuition.
At non-deterministic decision points, the LLM is invoked (via the $\mathsf{infer}$ operator) to guide symbolic reasoning: this is a guess of what transitions to perform, and can imply a guess of the final solution.
Say the transition system takes a sequence of steps that contradict the transitions guessed by the LLM.
The intuition for the steps that have been taken becomes part of the running intuition for the computation (via the $\mathsf{combine}$ operator), and thus acts as a counterexample to the LLM's previous guess when the $\mathsf{infer}$ operator is next invoked on the running intuition.
Example~\ref{ex:synthesis} demonstrates neural guessing and symbolic checking in an implementation of {\nsdm}-based program synthesis.

\paragraph{Better Computational Properties}

It is possible to implement {\nsdm}s that have good properties using techniques that are well established for symbolic algorithms.
For example, one could base an \nsdm on a symbolic transition system that has some particular property---say, it always terminates within $k$ steps.
The \nsdm would also terminate within $k$ steps, as the \nsdm can perform only transitions that are also possible in the underlying symbolic transition system.
Furthermore, when an \nsdm is implemented, the properties of the underlying transition system can be combined with a sensible instantiation of non-deterministic choice to ensure that the implementation has desirable properties (e.g., non-determinism is handled in a way that induces fair search, so that the overall algorithm can be a semi-decision procedure).

\section{\nsdm-Based Logic Programming}\label{sec:lang}

\begin{figure}
\centering

\begin{subfigure}{0.95\columnwidth}
  \begin{prompt}
    You will be asked questions about the following logic program: [\dots].
    We will use this program to [synthesize a loop invariant for the following C function: \dots].
    Please form some intuition about the scenario, and then guess a solution to the query [\dots] and a complete derivation for it, using the rules in the program.
  \end{prompt}
  \caption{Prompt for initializing intuition}\label{fig:prompta}
\end{subfigure}

\vspace{1em}

\begin{subfigure}{0.95\columnwidth}
  \begin{prompt}
  We have made the following conclusions: [\dots].
  This contradicts your guess in these ways: [\dots].
  Given our conclusions, update your guess for the solution to the query [\dots] and the derivation for that solution.
  \end{prompt}
  \caption{Prompt for updating intuition}\label{fig:promptb}
\end{subfigure}

\caption{The logic programming runtime interacts with the LLM to generate the initial intuition and extract guidance for symbolic transitions.
These prompts are illustrative; an implementation would require careful prompt engineering.}\label{fig:prompt}
\end{figure}

In this section, I describe a relatively naive implementation of the \nsdm model: a top-down logic programming language that calls an external LLM with intuition reified as text.
To build such a language, one could start with an abstract machine for top-down logic programming, such as a variation on the Warren Abstract Machine~\cite{Warren1983Abstract}.
As input, the abstract machine takes a query and a set of logical inference rules; the machine applies the inference rules to build a derivation tree for the query top down (i.e., starting at the query).
Because multiple rules might apply at the same point, evaluation is non-deterministic.
We assume that non-determinisim in the abstract machine can be instantiated in such a way as to provide fair search over the space of possible derivations.

The \nsdm runs the abstract logic programming machine on the given program and query, paired with initial intuition that can be formed from, say, the program text and context for the program (Figure~\ref{fig:prompta}).
The intuition includes a guess of the solution to the query, as well as a derivation for that query; when there are multiple candidate derivations to explore, this intuition can be used to guide decision making.

The \nsdm runs the abstract machine, updating the intuition state (via the $\mathsf{combine}$ operation) when the abstract machine makes a novel conclusion, which can potentially be used as a building block in a derivation.
If the abstract machine determines that it is unable to derive a conclusion used in the derivation guessed by the LLM, it updates the intuition with this fact, and uses the $\mathsf{infer}$ operation to prompt the LLM for an updated guess (Figure~\ref{fig:promptb}).
Hence, the \nsdm implements counterexample-guided guess-and-check logic.

Because LLM calls are expensive, an implementation of this \nsdm might choose to not invoke the LLM at every decision point.
For example, it might instead perform a fair logic programming search---every finite derivation is explored eventually---biased by the LLM's current guess for the derivation.
When the LLM updates its guess for the derivation, the bias of the logic programming search is also updated.
This algorithm would be a semi-decision procedure for any query with at least one finite successful derivation.

\begin{example}[\nsdm Program Synthesis]\label{ex:synthesis}
  Figure~\ref{fig:program} gives a fragment of enumerative program synthesis in \nsdm logic programming.
  Given the top-level query \lstinline|solution(X)|, the LLM would guess a value $v$ for the variable \lstinline|X|---i.e., the AST for a candidate program---and a derivation for the conclusion \lstinline|solution($v$)|.
  This conclusion is derived via one rule, which induces evaluation to produce a term through the subquery \lstinline|term(X)| and then check that the term satisfies a specification via the subquery \lstinline|verifies(X)|.
  The \lstinline|term| predicate defines all ASTs in the target language; evaluation enumerates through these ASTs, but is biased by the LLM's guessed derivation for the conclusion \lstinline|term($v$)|. 
  Similarly, the checking of a term via the \lstinline|verifies| predicate is biased by the LLM's guessed derivation for the conclusion \lstinline|verifies($v$)|.
  The LLM is prompted for new guesses if the logic it provided (i.e., the derivation) for generating or checking the term $v$ is proven wrong.
\end{example}

This approach leverages the impressive generative abilities of LLMs, as logic programming search should complete quickly if the LLM guesses correctly (or is close).
At the same time, the fair search strategy guarantees that symbolic reasoning still makes progress even when the LLM hallucinates and guesses incorrectly, and provides a suitable platform for implementing decision procedures and semi-decision procedures.
Outside the hassle of rewriting code as a logic program, it should be straightforward to port many LLM-powered AR algorithms from the sequential \nesy architecture to \nsdm logic programming:
where LLMs had been used as solution oracles, developers simply provide a predicate defining the grammar of candidate solutions (e.g., the \lstinline|term| predicate).

\paragraph{Experiments} In a proof-of-concept test, ChatGPT~\cite{openai2023gpt4} produced successful type-inference derivations for a non-algorithmic type system encoded as a logic program, providing some initial evidence that LLMs can generate accurate derivations for logic programs involving non-trivial choices.

\paragraph{Future Possibilities}

I have described a relatively naive implementation of the \nsdm model.
Other implementations are possible beyond logic programming.
Future work could investigate integrating an \nsdm more closely with a neural architecture, such that intuition is internal neural state rather than text.
It would be interesting to explore this integration not only at inference time, but also during model training.

\begin{figure}
\centering
  \begin{lstlisting}[language=prolog]
solution(X) :- term(X), verifies(X).
term(var(X)) :- is_var(X).
term(const(X)) :- is_const(X).
term(binop(Op, T1, T2)) :-
  op(Op), term(T1), term(T2). ...
  \end{lstlisting}
  \caption{A developer can implement LLM-based enumerative program synthesis by writing a purely symbolic program that is evaluated with an \nsdm-based language runtime.
  }\label{fig:program}
\end{figure}

\begin{acks}
  Thanks to Toby Murray for his encouragement to write a position paper, and Alice Feng for her encouragement to think bigger.
  Thanks to Nada Amin, William E. Byrd, Thanh Le-Cong, Toby Murray, Thanh-Dat Nguyen, and Karl Schoch for their insightful feedback on earlier versions of this paper.

\end{acks}

\bibliographystyle{ACM-Reference-Format}
\bibliography{main}

\end{document}